\documentclass[sigconf]{acmart}

\AtBeginDocument{%
  }

\usepackage{newfloat}
\usepackage{listings}

\usepackage{newfloat}
\usepackage{listings}

\usepackage{amsmath} 
\usepackage{caption} 
\usepackage{xspace}
\usepackage{multirow}
\usepackage{xcolor}
\usepackage{amsmath}
\usepackage{amsfonts}
\usepackage{bm}
\usepackage{soul}

\newcommand{\eg}{\emph{e.g.},\xspace}
\newcommand{\ie}{\emph{i.e.},\xspace}

\newcommand{\etc}{\emph{etc.}\xspace}

\newcommand{\eat}[1]{}
\newcommand{\hao}[1]{{\color{blue}{#1}}}

\newcommand{\TODO}[1]{{\color{red}TODO:{#1}}}

\usepackage[utf8]{inputenc} 
\usepackage[T1]{fontenc}    
\usepackage{booktabs}       
\usepackage{amsfonts}       
\usepackage{nicefrac}       
\usepackage{microtype}      
\usepackage{xcolor} 
\usepackage{graphicx}
\usepackage{subcaption}
\usepackage{enumitem}
\newcommand{\better}[1]{\textcolor{magenta}{#1}}
\usepackage{pifont}
\newcommand{\red}[1]{{\color{red}{#1}}}
\definecolor{deepgreen}{RGB}{0,128,0}
\newcommand{\green}[1]{{\color{deepgreen}{#1}}}

\setcopyright{acmlicensed}
\copyrightyear{2018}
\acmYear{2018}
\acmDOI{XXXXXXX.XXXXXXX}
\acmConference[Conference acronym 'XX]{Make sure to enter the correct
  conference title from your rights confirmation email}{June 03--05,
  2018}{Woodstock, NY}
\acmISBN{978-1-4503-XXXX-X/2018/06}

\begin{document}

\title{DAMBench: A Multi-Modal Benchmark  for Deep Learning-based Atmospheric Data Assimilation}

\author{Hao Wang}
\authornote{Both authors contributed equally to this research.}
\email{hwang574@connect.hkust-gz.edu.cn}
\affiliation{%
  \institution{The Hong Kong University of Science and Technology (Guangzhou)}
  \city{Guangzhou}
  \state{Guangdong}
  \country{China}
}

\author{Zixuan Weng}
\authornotemark[1]
\email{zxweng0701@gmail.com}
\affiliation{%
  \institution{The Hong Kong University of Science and Technology (Guangzhou)}
    \city{Guangzhou}
  \state{Guangdong}
  \country{China}
}

\author{Jindong Han}
\email{jhanao@connect.ust.hk}
\affiliation{%
  \institution{The Hong Kong University of Science and Technology}
    \city{Hong Kong SAR}
  \state{Hong Kong}
  \country{China}
}

\author{Wei Fan}
\email{weifan.oxford@gmail.com}
\affiliation{%
  \institution{The Hong Kong University of Science and Technology (Guangzhou)}
    \city{Guangzhou}
  \state{Guangdong}
  \country{China}
}

\author{Hao Liu}
\authornote{Corresponding Author}
\email{liuh@ust.hk}
\affiliation{%
  \institution{The Hong Kong University of Science and Technology (Guangzhou)}
    \city{Guangzhou}
  \state{Guangdong}
  \country{China}\\
  \institution{The Hong Kong University of Science and Technology}
    \city{Hong Kong SAR}
  \state{Hong Kong}
  \country{China}
}

\begin{abstract}
Data Assimilation is a cornerstone of atmospheric system modeling, tasked with reconstructing system states by integrating sparse, noisy observations with prior estimation.
While traditional approaches like variational and ensemble Kalman filtering have proven effective, recent advances in deep learning offer more scalable, efficient, and flexible alternatives better suited for complex, real-world data assimilation involving large-scale and multi-modal observations.
However, existing deep learning-based DA research suffers from two critical limitations: (1) reliance on oversimplified scenarios with synthetically perturbed observations, and (2) the absence of standardized benchmarks for fair model comparison.
To address these gaps, in this work, we introduce DAMBench, the first large-scale multi-modal benchmark designed to evaluate data-driven DA models under realistic atmospheric conditions. DAMBench integrates high-quality background states from state-of-the-art forecasting systems and real-world multi-modal observations~(\ie real-world weather stations and satellite imagery). 
All data are resampled to a common grid and temporally aligned to support systematic training, validation, and testing. 
We provide unified evaluation protocols and benchmark representative data assimilation approaches, including latent generative models and neural process frameworks. 
Additionally, we propose a lightweight multi-modal plugin to demonstrate how integrating realistic observations can enhance even simple baselines.
Through comprehensive experiments, DAMBench establishes a rigorous foundation for future research, promoting reproducibility, fair comparison, and extensibility to real-world multi-modal scenarios. 
Our dataset and code are publicly available at \url{https://github.com/figerhaowang/DAMBench}. 
\end{abstract}

\ccsdesc[500]{Computing methodologies~Artificial intelligence}
\ccsdesc[500]{Information systems~Geographic information systems}

\keywords{data assimilation, spatio-temporal data mining, AI for climate science}

\received{20 February 2007}
\received[revised]{12 March 2009}
\received[accepted]{5 June 2009}

\maketitle

\section{Introduction}
\begin{figure*}[ht]
    \centering
    \includegraphics[width=0.98\linewidth]{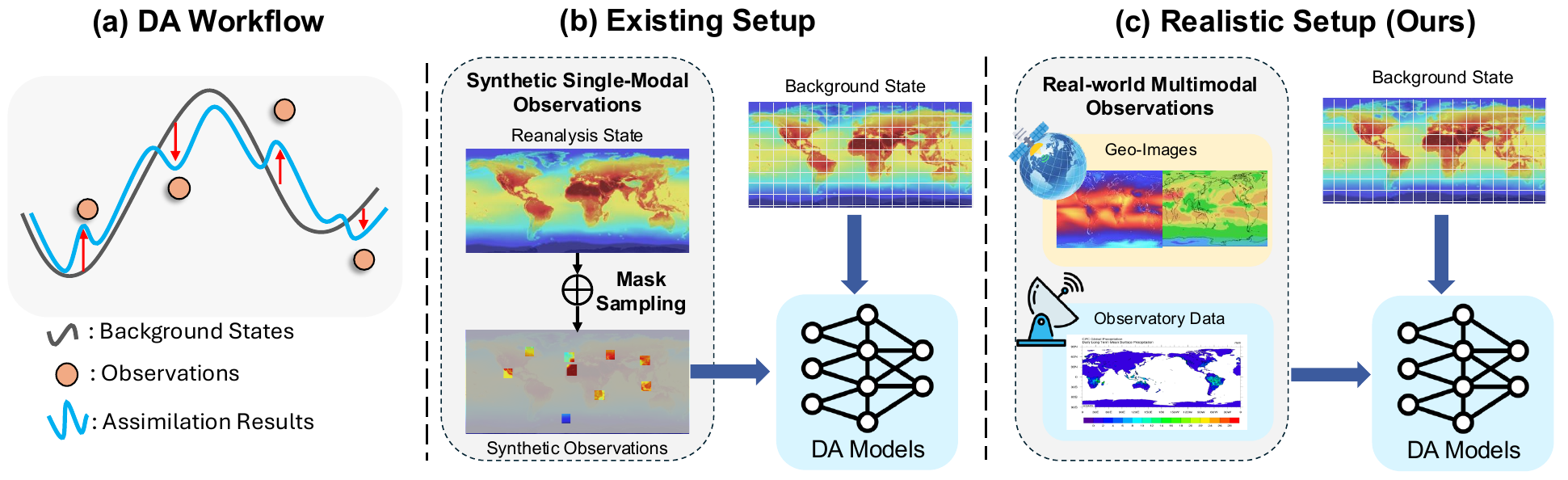}
    \vspace{-5pt}
    \caption{(a) DA recovers the system state (analysis states) by combining prior estimation (background states) with sparse and noisy observations (b) Existing methods often oversimplify the observation process by generating inputs via synthetic perturbations of background states, failing to capture the statistical and spatial structure of real-world measurements. (c) In contrast, our benchmark adopts realistic observational masks derived from reanalysis data and evaluates performance under operationally relevant conditions. 
    }
        \vspace{-5pt}
    \label{fig:intro_figure}
    \vspace{-5pt}
\end{figure*}

Data Assimilation~(DA) is fundamental to the atmospheric system, enabling the estimation of complex system states by combining heterogeneous observations with prior model estimation~\citep{lorenc1986analysis, evensen1994sequential, carrassi2018data}.
Traditional DA methods such as the Ensemble Kalman Filter (EnKF) and four-dimensional variational assimilation (4D-Var)~\citep{le1986variational, bannister2017review, courtier1998ecmwf} solve this as an optimization problem and reconciles physical models with noisy observations. 
However, these approaches rely on strong statistical assumptions and computationally expensive matrix operations, limiting their scalability for high-resolution, time-sensitive applications.
As the complexity and resolution of dynamic atmospheric system models increase, the need for more efficient, flexible, and data-adaptive DA strategies becomes increasingly urgent.

The rise of Deep Learning (DL) has revolutionized scientific modeling, spanning tasks such as weather forecasting~\cite{chen2023fengwu, lam2023graphcast}, ocean exploration~\citep{johnson2023oceanbench}, precipitation forecasting~\citep{schroeder2020rainbench}, \etc
DL-based systems like GraphCast~\citep{lam2023graphcast}, Pangu-Weather~\citep{bi2023pangu}, and FengWu~\citep{chen2023fengwu} have shown great superiority against traditional numerical weather prediction (NWP) models. 
Inspired by these successes, researchers have begun exploring DL-based methods for DA~\cite{huang2024diffda, rozet2023score, qu2024slam, chen2024fnp, chen2023adas}, leveraging its strengths in amortized inference over observations, scalable handling of high-dimensional data, and incorporating physical priors.
For example, DiffDA~\citep{huang2024diffda} frames DA as a conditional generative task using diffusion models, while SLAM~\citep{qu2024slam} introduces score-based diffusion in a shared latent space for multi-modal assimilation. In addition, neural process-based models such as FNP~\citep{chen2024fnp} further improve spatial flexibility and uncertainty quantification. Overall, these methods highlight the potential of Deep Learning in redefining the landscape of data assimilation.

Despite recent progress in deep learning-based DA, two key challenges hinder the systematic advancement of this field.
First, there is a notable lack of comprehensive and standardized benchmarking. Most existing studies evaluate newly proposed models in isolated experimental setups, with limited comparison with other representative methods. As a result, it remains challenging to assess the relative strengths, weaknesses, and generalization capabilities of different architectures under consistent conditions.
Second, the problem settings adopted in many existing studies are overly idealized and diverge from the operational requirements of DA. In real-world applications, DA systems ingest observations from multi-modal and heterogeneous sources, each exhibiting varying noise characteristics, spatial resolutions, and coverage. However, most DL-based DA methods to date simulate observations by perturbing background states with synthetic noise or filters, thus ignoring the inherent complexity and indirect nature of real observations. This oversimplification limits both the realism and practical applicability of such models in operational settings.

\begin{table*}[t]
\centering
\caption{Comparisons between DAMBench and other climate science benchmark datasets. Here, \green{\ding{51}} represents meeting a better standard, \red{\ding{55}} represents not meeting it. Our DAMBench is the first to support real-world DA with multi-modal observations.}
\resizebox{0.95\textwidth}{!}{
\begin{tabular}{cl|c|c|c|c|c|c|c}
\toprule
\multicolumn{2}{c|}{\multirow{2}{*}{\textbf{Evaluations}}} & \textbf{WeatherBench2} & \textbf{ChaosBench} & \textbf{OceanBench} & \textbf{RainBench}& \textbf{Terrra} & \textbf{DABench} & \textbf{DAMBench}  \\
& & \citep{rasp2024weatherbench} & \citep{nathaniel2024chaosbench} & \citep{johnson2023oceanbench} & \citep{schroeder2020rainbench}& \citep{chen2024terra} & \citep{wang2024dabench} & (Ours) \\
\hline
\multicolumn{2}{c|}{Year}&2024&2024&2023&2020&2024&2024&2025\\
\hline
Supported& Data Assimilation & \red{\ding{55}} & \red{\ding{55}}& \red{\ding{55}} &\red{\ding{55}} & \red{\ding{55}} & \green{\ding{51}} & \green{\ding{51}} \\

 Tasks& Weather Forecasting & \green{\ding{51}} & \green{\ding{51}} & \red{\ding{55}} & \green{\ding{51}} & \red{\ding{55}} & \red{\ding{55}} & \green{\ding{51}} \\
\hline

Multi-& Background State  & \green{\ding{51}} & \green{\ding{51}} & \red{\ding{55}} & \red{\ding{55}} & \red{\ding{55}} & \green{\ding{51}} & \green{\ding{51}} \\
Modality & Satellite Images  & \red{\ding{55}} & \red{\ding{55}} & \green{\ding{51}} & \green{\ding{51}} & \green{\ding{51}} & \red{\ding{55}} & \green{\ding{51}} \\
& Observatory Data & \red{\ding{55}} & \red{\ding{55}} & \red{\ding{55}} & \red{\ding{55}} & \red{\ding{55}} & \red{\ding{55}} & \green{\ding{51}} \\
\hline

Evaluation & Numerical Metrics & \green{\ding{51}} & \green{\ding{51}} & \green{\ding{51}} & \green{\ding{51}} & \green{\ding{51}} & \green{\ding{51}} & \green{\ding{51}} \\
Metrics & Physical-Based Metrics & \red{\ding{55}} & \green{\ding{51}} & \green{\ding{51}} & \red{\ding{55}}  & \red{\ding{55}} & \red{\ding{55}} & \green{\ding{51}} \\
\bottomrule
\end{tabular}}
\label{tab:benchmark-comparison}
\end{table*}

To bridge these gaps, we introduce DAMBench, a comprehensive, multi-modal benchmark specifically designed for DL-based data assimilation in atmospheric systems. 
Unlike prior works that rely on synthetic setups or simplified pseudo-observations, DAMBench emphasizes the assimilation of real-world multi-modal observations.
It enables systematic comparison of state-of-the-art DA models under a standardized evaluation framework.
DAMBench consists of three key data components: (i)background states derived from ERA5 reanalysis, a physically consistent reconstruction of historical weather states produced by the European Centre for Medium-Range Weather Forecasts (ECMWF); 
(ii) station-based observations collected from global weather stations~\citep{chen2008assessing}. We use global precipitation data which captures complex inter-action on land, ocean, and atmosphere, provided by the Climate Prediction Center; and 
(iii) satellite-based observations in the form of rasterized satellite imagery called outgoing longwave radiation (OLR) data collected on-board the NOAA-14, NOAA-16, and NOAA-18 satellites, which is an important parameter that is closely related to the planetary energy budget where temperature profile is coupled.
All data are temporally aligned to support daily assimilation cycles. 
We evaluate representative DL-based DA approaches under both uni-modal and multi-modal settings. 
In particular, we design a lightweight plugin named multi-modal representation adapter as a testbed to assess how the integration of multi-modal observations can enhance existing DA models. 
Experimental results reveal that even this simple plugin can significantly boost model performance when real-world multi-modal data are leveraged, highlighting the critical need for benchmarks grounded in authentic observation regimes.
Our contributions are summarized as follows:
\begin{itemize}[leftmargin=*, topsep=5pt, partopsep=5pt, itemsep=3pt]
  \item We propose \textbf{DAMBench}, the first multi-modal, large-scale benchmark tailored for evaluating deep learning-based data assimilation methods in a realistic, operationally grounded setting.
  \item We demonstrate \textbf{the necessity of real-world multi-modal observations} in DA and explore how to leverage these observations to boost existing approaches effectively.
  \item Our DAMBench provides standardized protocols and conducts extensive experiments on existing DL-based DA approaches, enabling reproducible and fair assessment of existing and future DA approaches.
\end{itemize}

\section{Related Work}

\subsection{DL Methods for Data Assimilation}

The application of deep learning (DL) to data assimilation (DA) has progressed rapidly in recent years, particularly in tandem with the rise of large weather models (LWMs) such as GraphCast~\citep{lam2023graphcast}, FengWu~\citep{chen2023fengwu}, GenCast~\citep{gencast}, and Pangu-Weather~\citep{bi2023pangu}. These models have demonstrated that DL-based systems trained on reanalysis data can match or exceed the performance of traditional numerical weather prediction (NWP) frameworks, offering improved scalability and reduced computational cost.

Building on these advances, a number of studies have proposed deep learning-based DA methods aimed at recovering latent atmospheric states from sparse and noisy observations~\citep{ensf,chen2024fnp,xiao2025vae,rozet2023score}. Notably, neural process-based models like FNP~\citep{chen2024fnp} leverage resolution-agnostic architectures to assimilate observations of varying fidelity. VAE-Var~\citep{xiao2025vae} employs variational autoencoders~\citep{kingma2013auto} to model background uncertainty in a probabilistic manner. Among generative methods, score-based and diffusion models have gained increasing attention. For example, SDA~\citep{rozet2023score} utilizes reverse-time stochastic differential equations to infer high-dimensional posteriors, and has been extended to geophysical systems such as two-layer quasi-geostrophic flows~\citep{rozet2023scoretwo}. SLAM~\citep{qu2024slam} extends the SDA model with conditional information obtained from various observations.  DiffDA~\citep{huang2024diffda} integrates GraphCast-based forecasting with a diffusion-based posterior correction framework, demonstrating promising performance under pseudo-observation settings.

While these methods highlight the potential of DL for DA, most studies are conducted using custom data pipelines and observation configurations, which impedes reproducibility and fair comparison. Differences in model inputs, observation modalities, and evaluation protocols make it difficult to assess the relative effectiveness of different approaches. As a result, there is a growing consensus that standardized benchmarks are essential for enabling systematic progress and robust evaluation in deep learning-based data assimilation.

\subsection{DL Benchmarks for Climate Science}

Benchmarks have been instrumental in driving progress in deep learning for climate science, particularly in the context of deterministic weather and ocean forecasting. Early efforts such as WeatherBench2~\cite{rasp2024weatherbench} and ChaosBench~\cite{nathaniel2024chaosbench} established unified experimental protocols and evaluation metrics based on reanalysis data, enabling reproducible model development and systematic comparison across different spatial resolutions and forecast horizons. These benchmarks have significantly lowered the barrier for entry into climate forecasting research and catalyzed rapid model innovation.
OceanBench~\cite{johnson2023oceanbench} further extended this paradigm to the ocean domain by providing a modular and reproducible evaluation suite for sea surface height reconstruction and reanalysis correction. Importantly, OceanBench incorporated realistic observational datasets from satellite altimetry and numerical ocean simulations, establishing a multi-modal framework for learning-based geophysical data recovery. This marked an important step toward bridging the gap between academic research and operational requirements.


Despite these advances, existing benchmarks remain largely tailored to forward prediction tasks and are not designed to evaluate the unique challenges of DA. They lack structured triplets of background states, observations, and analysis targets; do not support posterior inference objectives; and offer limited tools for integrating heterogeneous data sources. In operational settings, however, DA systems commonly incorporate sparse, noisy, and modality-diverse observations from satellites, ground stations, radar, or other sensors. Yet, current DL-based DA studies often resort to ad hoc setups, replacing real observations with perturbed or filtered background fields. This oversimplification neglects the indirect, noisy nature of real-world measurements, hindering model generalization and limiting practical applicability.
While DABench~\cite{wang2024dabench} represents a promising first attempt at building a dedicated benchmark for deep learning-based data assimilation, it is limited in scope. DABench primarily focuses on single-modal inputs and includes only a narrow selection of baseline methods, lacking support for real-world multi-modal observations and posterior generative modeling. Consequently, it does not fully reflect the complexity or operational relevance of modern DA systems.

To date, no benchmark provides structured, multi-modal DA tasks at global atmospheric scales with comprehensive support for posterior modeling, uncertainty quantification, and multi-modal observational fusion. These gaps highlight the urgent need for a unified benchmark that enables realistic and reproducible evaluation of deep learning-based DA methods. To address this, we introduce \textbf{DAMBench}, a multi-modal, real-world and operationally grounded benchmark tailored to the demands of atmospheric data assimilation.

\section{Preliminary}

In this section, we formalize the data assimilation (DA) problem under a probabilistic framework and describe the mathematical structure underlying real-world atmospheric system modeling. This formulation serves as the foundation for benchmarking and evaluating learning-based DA algorithms within our proposed DAMBench.

Let $\mathbf{x}(t) \in \mathbb{R}^m$ denote the latent state of a high-dimensional dynamical system (e.g., the global atmosphere) at time $t$. The system evolves according to a (possibly nonlinear) physical transition operator $\mathcal{M}_{t-1 \rightarrow t}: \mathbb{R}^m \rightarrow \mathbb{R}^m$, such that:
\begin{equation}
    \mathbf{x}(t) = \mathcal{M}_{t-1 \rightarrow t}(\mathbf{x}(t-1)),
\end{equation}
where $\mathcal{M}$ represents the underlying physical dynamics, often approximated by numerical weather prediction (NWP) models. In practice, the true state $\mathbf{x}(t)$ is not directly observable. Instead, we obtain partial and noisy observations $\mathbf{y}(t) \in \mathbb{R}^n$ through an observation operator $\mathcal{H}: \mathbb{R}^m \rightarrow \mathbb{R}^n$:
\begin{equation}
    \mathbf{y}(t) = \mathcal{H}(\mathbf{x}(t)) + \boldsymbol{\epsilon}(t),
\end{equation}
where $\boldsymbol{\epsilon}(t)$ is typically modeled as Gaussian noise: $\boldsymbol{\epsilon}(t) \sim \mathcal{N}(0, \mathbf{R})$, with $\mathbf{R}$ being the observational error covariance.

The goal of data assimilation is to estimate the posterior distribution of the system state given the sequence of observations up to time $t$:
\begin{equation}
    p(\mathbf{x}(t) \mid \mathbf{y}_{1:t}) \propto p(\mathbf{y}(t) \mid \mathbf{x}(t)) \cdot p(\mathbf{x}(t) \mid \mathbf{y}_{1:t-1}),
\end{equation}
where the prior $p(\mathbf{x}(t) \mid \mathbf{y}_{1:t-1})$, commonly referred to as the \textit{background state} $\mathbf{x}_b(t)$, is obtained by propagating the previous analysis forward through the system dynamics:
\begin{equation}
    \mathbf{x}_b(t) = \mathcal{M}_{t-1 \rightarrow t}(\hat{\mathbf{x}}(t-1)),
\end{equation}
with $\hat{\mathbf{x}}(t-1)$ representing the posterior estimate from the previous assimilation step.

In classical variational and Bayesian DA frameworks, the posterior is typically approximated by minimizing the negative log-posterior, resulting in the following surrogate objective:
\begin{equation}
\begin{split}
\mathcal{L}_{\text{DA}}(\mathbf{x}) =\ 
& \frac{1}{2} (\mathbf{x} - \mathbf{x}_b)^\top \mathbf{B}^{-1} (\mathbf{x} - \mathbf{x}_b) + \\
& \frac{1}{2} (\mathbf{y} - \mathcal{H}(\mathbf{x}))^\top \mathbf{R}^{-1} (\mathbf{y} - \mathcal{H}(\mathbf{x})),
\end{split}
\end{equation}
where $\mathbf{B}$ is the background error covariance matrix. The first term penalizes deviations from the forecast prior, while the second enforces consistency with observed data. Solving this optimization yields the best linear unbiased estimate under Gaussian assumptions.

In deep learning-based DA, the inference process is instead approximated using a parametric model $f_\theta$, typically realized as a neural network, which directly maps the background and observation history to an analysis estimate:
\begin{equation}
    \hat{\mathbf{x}}(t) = f_\theta(\mathbf{y}_{1:t}, \mathbf{x}_b(t)),
\end{equation}
thereby bypassing explicit optimization and enabling efficient amortized inference. Depending on the formulation, this may be posed as a supervised regression task, a generative modeling problem (e.g., using diffusion models or VAEs), or a structured latent process approximation.

In this work, we adopt the state-of-the-art deep forecasting model FengWu~\cite{chen2023fengwu} as a surrogate for the physical simulator $\mathcal{M}$, providing high-quality, multivariate background fields. This enables a fully data-driven DA framework, facilitating large-scale integration of real-world, multi-modal observations with forecast priors in a realistic and reproducible setting.

\begin{table*}[ht]
\centering
\caption{List of atmospheric variables contained in the reanalysis data in DAMBench. }
\label{tab:variables}
\begin{tabular}{ll|ll}
\toprule
\textbf{Upper-air variables (Short name \& Unit)} & \textbf{Levels} & \textbf{Surface variables/Constants (Short name \& Unit)} & \textbf{Levels}\\
\midrule
Geopotential (z \& $m^2s^{-2}$) & 13 & Temperature at 2m height (t2m \& K) & single\\
Temperature (t \& K) & 13 & x-direction wind at 10m height (u10 \& $ms^{-1}$) & single\\
Specific\_humidity (q \& $kgkg^{-1}$) & 13 & y-direction wind at 10m height (v10 \& $ms^{-1}$) & single \\
x-direction wind (u \& $ms^{-1}$) & 13 & mean\_sea\_level\_pressure (msl \& $Pa$) & single\\
y-direction wind (v \& $ms^{-1}$) & 13 &  &  \\
\bottomrule
\end{tabular}
\end{table*}

\section{Benchmark}
To rigorously evaluate learning-based data assimilation (DA) algorithms, DAMBench introduces a standardized benchmark suite that mirrors the structure of operational DA systems. Each task instance consists of three components: a noisy and partial observation $\mathbf{y}(t)$, a background prior $\mathbf{x}_b(t)$, and a reference analysis $\mathbf{x}^\ast(t)$ derived from reanalysis. Together, these triplets define a supervised DA setting where the goal is to estimate $\mathbf{x}^\ast(t)$ given $(\mathbf{y}(t), \mathbf{x}_b(t))$.

While prior work often adopts simplified or synthetic DA settings—such as perturbing background fields with noise—DAMBench is designed to support realistic uni-modal and multi-modal configurations. In this section, we describe the uni-modal benchmark setup, which focuses on assimilating single-source, observation-masked reanalysis data, and defer the multi-modal extensions to the next section.

\subsection{Benchmark: Data Composition and Design}
\label{sec:uni-data}

In this section, we introduce the composition of our dataset in uni-modal benchmark, using masked reanalysis data as observations. 

\subsubsection{Reanalysis Data}
We adopt ERA5 reanalysis data~\cite{hersbach2019era5} as the ground-truth approximation of the true atmospheric state. ERA5 is generated using a state-of-the-art four-dimensional variational (4D-Var)~\citep{bannister2017review, courtier1998ecmwf} DA system operated by ECMWF, assimilating a wide range of global observations over the past decades. It serves as the target variable $\mathbf{x}^\ast(t)$ in our DA setup, providing temporally and spatially consistent fields for supervised learning. 

 As can be seen in Table~\ref{tab:variables}, we choose to conduct experiments on a total of 69 variables, including five upper-air variables with 13 pressure levels, and four surface variables:
\begin{itemize}[leftmargin=*, topsep=0pt, partopsep=0pt]
    \item \textbf{Upper-air variables} (at 13 pressure levels): geopotential (z), temperature (t), specific humidity (q), x-direction wind (u), and y-direction wind (v). These form 65 pressure-level-specific channels, e.g., z500, t850, \etc
    \item \textbf{Surface variables}: 10m wind components (u10, v10), 2m temperature (t2m), and mean sea-level pressure (msl).
\end{itemize}
We extract data from 2020–2024 for training and validation, ensuring wide temporal coverage and high data diversity. All fields are resampled to a $121 \times 240$ grid for efficiency and consistency.

\subsubsection{Background Data}
Background states $\mathbf{x}_b(t)$ are generated by running pretrained numerical or machine learning-based forecasting models forward in time, starting from previously assimilated analyses. These fields act as priors in the Bayesian formulation of DA and are typically biased or noisy due to imperfect models or chaotic divergence. In DAMBench, we include background fields deep learning model FengWu~\cite{chen2023fengwu} to generate background states $\mathbf{x}_b(t)$, serving as priors in the assimilation task.  Formally, the goal of the assimilation model is to refine $\mathbf{x}_b(t)$ into an improved estimate $\hat{\mathbf{x}}(t)$ that is consistent with observations.

\subsubsection{Observation Data}
Observations $\mathbf{y}(t)$ serve as the conditioning input to the DA model and reflect the noisy, partial information available at each timestep. 
To emulate realistic observation settings while ensuring full supervision during training, DAMBench generates observation data $\mathbf{y}(t)$ by applying a spatial sampling mask over the reanalysis fields $\mathbf{x}^*(t)$:
\begin{equation}
    \mathbf{y}(t) = \mathbf{M}(t) \odot \mathbf{x}^*(t) + \boldsymbol{\epsilon}(t),
\end{equation}
where $\mathbf{M}(t) \in \{0,1\}^m$ is a binary mask that determines which spatial grid points are observed at time $t$, $\odot$ denotes element-wise multiplication, and $\boldsymbol{\epsilon}(t) \sim \mathcal{N}(0, \mathbf{R})$ is additive Gaussian noise simulating sensor errors.

This masking operator $\mathcal{H}$ implicitly reflects the observation geometry of real-world sensing systems, such as sparse station measurements or incomplete satellite swaths. By varying the sparsity, location, and modality-specific noise levels in $\mathbf{M}(t)$, DAMBench enables controlled experiments under different observation regimes.


In DAMBench, a single DA instance is defined as a tuple 
\begin{align}
    (\mathbf{x}_b(t), \mathbf{y}(t)) \mapsto \mathbf{x}^\ast(t), 
\end{align}
where the model aims to reconstruct the reanalysis state $\mathbf{x}^\ast(t)$ given the background $\mathbf{x}_b(t)$ and observations $\mathbf{y}(t)$. This setup is consistent with classical DA pipelines and enables supervised training and evaluation of learning-based DA models. Crucially, the diversity of observations, multiple background sources, and high-quality reanalysis fields allow DAMBench to support both uni-modal and multi-modal DA settings, as well as single-step or sequential assimilation tasks.

\begin{table*}[ht]
  \caption{Quantitative performance comparison for deep learning methods for data assimilation. The best performance are shown in \textbf{bold} while the second best is \underline{underscored}. The baseline results are average of 5 parallel experiments. We show the SpecDiv, MSE($10^{-2}$) and MAE$(10^{-2})$ over all variables and RMSE of part of the variables. }
  \label{tab:main_experiments}
  \centering
  \begin{tabular}{l|c|cc|cccccccc}
    \toprule
    \multirow{2}{*}{Model} & \multirow{2}{*}{SpecDiv $\downarrow$} & \multirow{2}{*}{MSE$\downarrow$} & \multirow{2}{*}{MAE$\downarrow$} & \multicolumn{8}{c}{RMSE$\downarrow$} \\
    & & & & z500 & t850 & t2m & u10 & v10 & u500 & v500 & q700 ($10^{-4}$) \\
    \midrule
    Background & 0.153 & 2.88 & 8.61 & 45.455 & 0.7200 & 0.7790 & 0.9336 & 0.9645 & 1.7278 & 1.7535 & 6.7220 \\
    \midrule
    \midrule

    Adas \citep{chen2023adas} & 0.116 & \underline{2.31} & 7.65 & 30.100 & \textbf{0.6750} & 0.7350 & 0.8400 & 0.8600 & 1.4950 & 1.4900 & 6.5400 \\
    ConvCNP \citep{gordon2019convcnp} & 0.125 & 2.49 & 7.98 & 31.253 & \underline{0.6944} & 0.7662 & 0.8334 & 0.8553 & 1.5770 & 1.5876 & 6.5717 \\ 
    FNP \citep{chen2024fnp} & \underline{0.063} & \textbf{2.30} & \textbf{7.54} & \underline{28.500} & 0.6985 & \underline{0.7100} & \underline{0.7650} & \textbf{0.7650} & \textbf{1.4350} & \underline{1.4600} & \underline{6.4698} \\
    VAE-VAR \citep{xiao2025vae} & \textbf{0.052} & \underline{2.31} & \underline{7.60} & \textbf{27.000} & 0.6970 & \textbf{0.7050} & \textbf{0.7560} & \underline{0.7770} & \underline{1.4500} & \textbf{1.4500} & 6.4700 \\
    SDA \citep{rozet2023score} & 0.117 & 2.65 & 8.02 & 38.000 & 0.7100 & 0.7500 & 0.8800 & 0.9100 & 1.6500 & 1.7000 & 6.6100 \\
    SLAM \citep{qu2024deep} & {0.091} & 2.55 & 7.94 & 32.500 & 0.7020 & 0.7300 & 0.8000 & 0.7800 & 1.5000 & 1.4700 & \textbf{6.4690} \\

    \bottomrule
  \end{tabular}
\end{table*}

\subsection{Overview of the Existing Baselines}
Recent advances in data-driven data assimilation (DA) have led to the development of diverse frameworks that leverage deep learning to bridge observational data and model forecasts. These methods aim to address the limitations of traditional assimilation schemes by introducing neural architectures capable of capturing complex spatial-temporal dependencies, handling irregular observation patterns, and modeling uncertainty.
In this section, we provide an overview of representative approaches in this line of research. Each method offers unique design philosophies and technical innovations, contributing to the evolving landscape of DL-augmented assimilation systems.
\begin{itemize}[leftmargin=*, topsep=5pt, partopsep=0pt]
    
\item  Adas \citep{chen2023adas} represents a significant step toward end-to-end DL-based numerical weather prediction. It incorporates domain knowledge from traditional DA (e.g., confidence weighting) into a gated neural network architecture. Adas employs gated convolutions and gated cross-attention, modulated by a confidence matrix, to fuse sparse observations with background states. 

\item  ConvCNP \citep{gordon2019convcnp} belongs to the neural process family and introduces translation-equivariant functional embeddings to support structured prediction over spatial domains. It models the prediction function as a continuous convolutional map over a discretized input space, enabling flexible and uncertainty-aware interpolation across irregular or sparse observation contexts. 

\item  FNP \citep{chen2024fnp} is designed to overcome the resolution limitations in traditional AI-based data assimilation frameworks. It builds on neural processes by integrating Fourier-based embeddings to extract spatial patterns across varying grid scales. 

\item  VAE-Var \citep{xiao2025vae} blends deep generative modeling with classical variational data assimilation. It introduces a variational autoencoder to learn a more expressive, non-Gaussian background distribution $p(x \mid x_b)$, alleviating the strong Gaussian assumption in conventional 3D-Var and 4D-Var methods. 

\item  SDA \citep{rozet2023score} proposes a principled, generative approach to trajectory-level data assimilation by leveraging continuous-time score-based diffusion. 
A score model is trained on short trajectory segments, enabling the non-autoregressive generation of full state trajectories via reverse SDE sampling.

\item  SLAMS \citep{qu2024deep} is a diffusion-based data assimilation framework designed to operate in a unified latent space. 
It embeds multi-modal observations (e.g., satellite images, in-situ sensor readings) and background states into a shared latent representation, enabling conditional generation via score-based diffusion.

\end{itemize}

\subsection{Overview of Metrics}

Referring to Pangu-Weather~\citep{bi2023pangu} and FengWu~\citep{chen2023fengwu}, we chosoe the latitude-weighted root-mean-square error (RMSE) as our primary metric. Given the estimate $\hat{x}_{h,w,c}$ and its ground truth $x_{h,w,c}$ for the $c$-th channel, the RMSE is defined as:
\begin{equation*}
\resizebox{0.98\linewidth}{!}{$
\operatorname{RMSE}(c) = 
\sqrt{\frac{1}{H\cdot W}\sum\nolimits_{h,w} H \frac{\operatorname{cos}(\alpha_{h,w})}{\sum_{h'=1}^{H} \operatorname{cos}(\alpha_{h',w})}(x_{h,w,c} - \hat{x}_{h,w,c})^{2}}
$}
\end{equation*}
where $H$ and $W$ represent the number of grid points in the longitudinal and latitudinal directions, respectively, $C$ denotes the number of variable channels and $\alpha_{h,w}$ is the latitude of point $(h,w)$.
Moreover, we consider the common-used numerical metrics mean average error (MAE) and mean square error (MSE), defined as follow:
\begin{equation}
\operatorname{MSE} = \frac{1}{H \cdot W \cdot C} \sum\nolimits_{h,w,c}   (x_{h,w,c} - \hat{x}_{h,w,c})^2
\end{equation}
\begin{equation}
\operatorname{MAE} = \frac{1}{H \cdot W \cdot C } \sum\nolimits_{h,w,c} \left| x_{h,w,c} - \hat{x}_{h,w,c} \right|
\end{equation}
Moreover, to evaluate the physical coherence of the generated state, we leverage the physics-based metric, Spectral Divergence (SpecDiv)~\citep{nathaniel2024chaosbench, wang2025phyda} that measures the deviation between the power spectra of assimilated state and ground truth.  
SpecDiv follows principles from Kullback–Leibler (KL) divergence where we compute the expectation of the log ratio between target $S^\prime(\omega)$ and prediction $\hat{S}^\prime(\omega)$ spectra, and is defined as:
\begin{equation}
    \mathcal{M}_{SpecDiv} = \sum_{\omega} S^{\prime}(\omega) \cdot \log(S^{\prime}(\omega) / \hat{S}^{\prime}(\omega)),
    \label{eq:specdiv}
\end{equation}
where $\omega \in \mathbf{\Omega}$, and $\mathbf{\Omega}$ is the set of all scalar wavenumbers from 2D Fourier transform. More details can be seen in Appendix.

\subsection{Experiments Analysis}
In this part, we present the experiments conducted on the DAMBench  where all models were trained using a fair and comprehensive pipeline. However, due to space limitations, it is not possible to discuss all the results in detail within the confines of this paper. 
Following the setting of previous works~\citep{chen2023fengwu, chen2024fnp, chen2023adas}, we show the MSE, MAE and SpecDiv over all variables and RMSE for part of the variables.
These experiments demonstrate good performance for all the existing DL models.

\subsubsection{Experiment Setups}

We conduct a thorough and reproducible evaluation of state-of-the-art (SOTA) deep learning-based data assimilation (DA) methods under a unified experimental framework. All models are trained and evaluated using the dataset described in Section~\ref{sec:uni-data}. Specifically, the training set spans the years 2000–2022, the year 2023 is used for validation, and the year 2024 is reserved for final testing. All models ingest background fields generated via a 24-hour forecast using the Fengwu model~\citep{chen2023fengwu}, initialized from ERA5 reanalysis states. Observations are simulated by applying a binary mask to the reanalysis fields, retaining only 5\% of the grid points. We train all models using the AdamW optimizer~\citep{adamw}. The learning rate is scheduled with a warm-up cosine annealing strategy to stabilize early training dynamics and promote convergence. All experiments are conducted on four NVIDIA A800 GPUs. 

For fair comparison, all baseline implementations are adapted from the official open-source repositories of the respective methods, with hyperparameters and architectural configurations preserved as originally reported. All experiments are executed under a fixed random seed to ensure consistent initialization and data shuffling, enabling reproducible outcomes. Unless otherwise specified, all reported metrics are averaged over five independent runs to mitigate the effects of training stochasticity.

\subsubsection{Model Performance}
We evaluate the performance of several representative deep learning-based data assimilation (DA) methods using the proposed DAMBench. Table~\ref{tab:main_experiments} summarizes quantitative results across four metrics: SpecDiv, MSE, MAE and latitude-weighted RMSE, computed over a full one-year assimilation cycle. Evaluation covers 69 atmospheric variables, with detailed RMSE statistics reported for a subset of representative variables: z500, t850, t2m, u10, v10, u500, v500, and q700.

All DA models demonstrate substantial improvements over the background forecast provided by FengWu~\citep{chen2023fengwu}. 
Among all approaches, FNP~\citep{chen2024fnp} consistently achieves the best or second-best performance across all metrics. 
VAE-VAR~\citep{xiao2025vae} also delivers strong results, especially for near-surface variables like t2m and t850. 
SDA~\citep{rozet2023score} and SLAM~\citep{qu2024slam} trail behind other methods in overall MSE and MAE. While these score-based generative approaches have shown promise in synthetic or low-dimensional setups, they face limitations in large-scale DA tasks. 
These results collectively underscore the strengths and limitations of current deep DA paradigms.  The large variation in performance across variables (e.g., surface vs upper-air) highlights the importance of multi-resolution and modality-aware design, which we further explore in the multi-modal extensions presented in the following section.

\begin{figure}[t]
    \centering
    \includegraphics[width=0.95\linewidth]{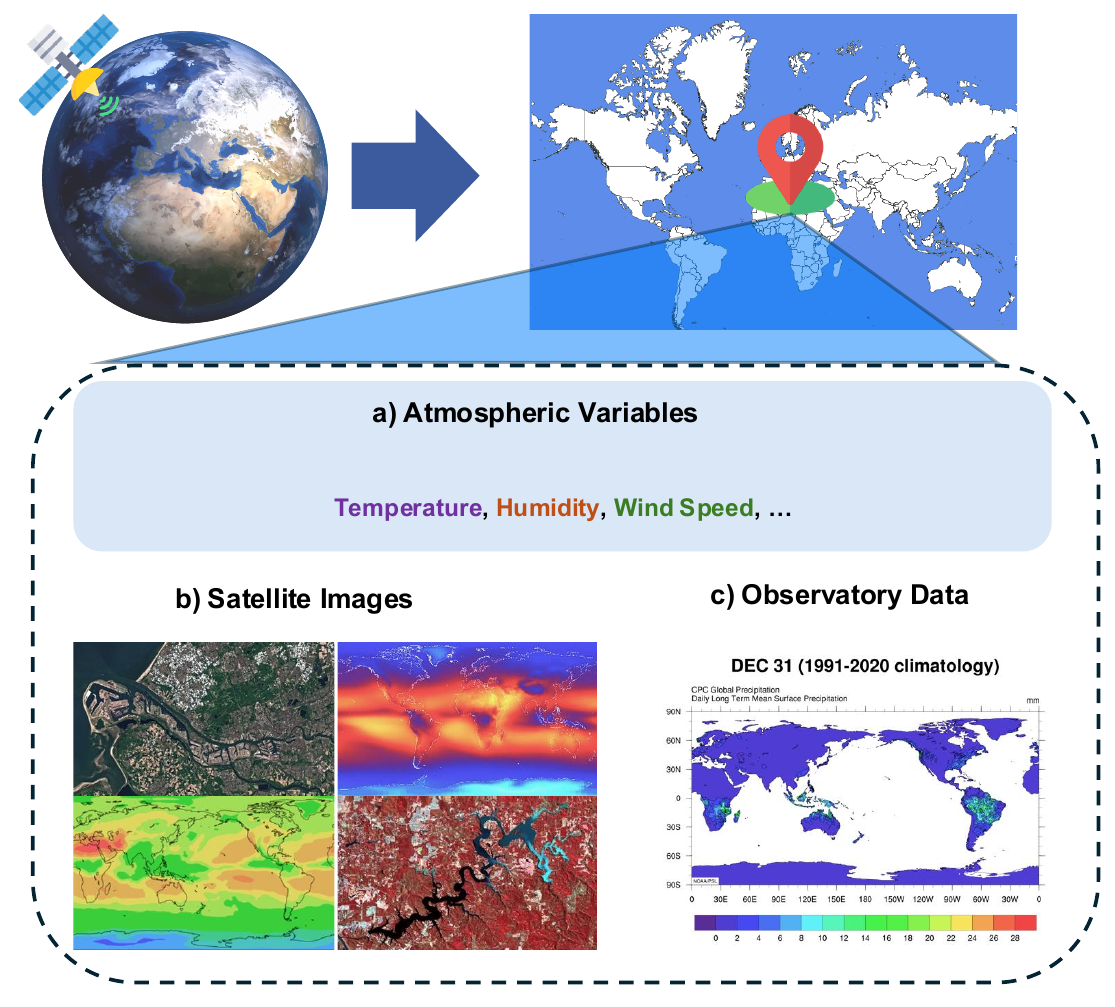}
    \caption{Overview of DAMBench's real-world observations with multi-modal information. We collected multi-modal real-world observations from both satellites and observatories, enabling more realistic exploration towards deep learning-based data assimilation.}
    \label{fig:observation_components}
    \vspace{-10pt}
\end{figure}
\section{Rethinking the Effectiveness of Multi-Modal Observations}

\begin{table*}[htbp]
  \caption{ Quantitative performance comparison for deep learning methods for data assimilation with additional \textit{multi-modal information}. The model named "Multi-" represents the model with multi-modal observation. The experiment results are average of 5 parallel experiments. We show the MSE($10^{-2}$) and MAE($10^{-2}$) over all variables and RMSE of part of the variables. And Imp represents the MSE improvement of the baseline models with multi-modal observations. }
  \label{tab:multimodal_experiments}
  \centering
  \resizebox{0.98\linewidth}{!}{
  \begin{tabular}{l|c|cc|cccccccc|c}
    \toprule
    \multirow{2}{*}{Model} & \multirow{2}{*}{SpecDiv $\downarrow$} & \multirow{2}{*}{MSE$\downarrow$} & \multirow{2}{*}{MAE$\downarrow$} & \multicolumn{8}{c}{RMSE$\downarrow$} & \multirow{2}{*}{Imp $\uparrow$}\\
    & & & & z500 & t850 & t2m & u10 & v10 & u500 & v500 & q700 ($10^{-4}$) &\\
    \midrule
    Background & 0.153 & 2.88 & 8.61 & 45.455 & 0.7200 & 0.7790 & 0.9336 & 0.9645 & 1.7278 & 1.7535 & 6.7220 &-\\
    \midrule
    \midrule

    Adas \citep{chen2023adas} &{0.116 } & 2.31 & 7.65 & 30.100 & {0.6750} & 0.7350 & 0.8400 & 0.8600 & 1.4950 & 1.4900 & 6.5400 &-\\
    Multi-Adas & 0.060 & {2.20} & {7.30} & 27.800 & 0.6700 & {0.6900} & {0.7400} & {0.7400} & {1.4000} & {1.4200} & {6.3500} & 4.35\%\\
    \hline

    ConvCNP \citep{gordon2019convcnp} &{0.125 } & 2.49 & 7.98 & 31.253 & 0.6944 & 0.7662 & 0.8334 & 0.8553 & 1.5770 & 1.5876 & 6.5717 &-\\ 
    Multi-ConvCNP & 0.123 & 2.44 & 7.82 & 30.628 & 0.6805 & 0.7510 & 0.8170 & 0.8380 & 1.5750 & 1.5560 & 6.5400 & 2.01\%\\
    \hline

    FNP \citep{chen2024fnp}&{0.063  }& {2.30}  & {7.54} & 28.500 & 0.6985 & 0.7100 & 0.7650 & {0.7650} & {1.4350} & 1.4600 & {6.4698} &-\\
    Multi-FNP & {0.059} & {2.16} & {7.09} & {26.790} & {0.6566} & {0.6674} & {0.7191} & {0.7191} & {1.3489} & {1.3724} & {6.0800} & 6.09\%\\
    \hline

    VAE-VAR \citep{xiao2025vae} &{0.052}  & 2.31 & 7.60 & {27.000} & 0.6970 & {0.7050} & {0.7560} & 0.7770 & 1.4500 & {1.4500} & 6.4700 &-\\
    Multi-VAE-VAR & {0.048} & {2.13} & {6.99} & {24.840} & {0.6412} & {0.6486} & {0.6955} & {0.7148} & {1.3340} & {1.3340} & {5.9500} & 7.79\%\\
    \hline

    SDA  \citep{rozet2023score}   &0.117   & 2.65 & 8.02 & 38.000 & 0.7100 & 0.7500 & 0.8800 & 0.9100 & 1.6500 & 1.7000 & 6.6100 &-\\
    SLAM  \citep{qu2024deep}  &0.091  & 2.55 & 7.94 & 32.500 & 0.7020 & 0.7300 & 0.8000 & 0.7800 & 1.5000 & 1.4700 & 6.5000 &3.77\%\\

    \bottomrule
  \end{tabular}
  }
\end{table*}

With the recent advances of multi-modal deep learning modeling, the natural progression is the integration of these deep learning principles into data assimilation for improved state estimation, unresolved scale (parameterization) inference, and parameter inference, either separately or in combination.
In particular, the rapid adoption of controllable deep generative modeling in text-conditioned image generation or image-controlled video synthesis could pave the way toward a robust deep learning data assimilation framework.

However, previous works on deep learning data assimilation usually used simplified scenarios: they tend to only assimilate perturbed states as pseudo-observations through filtering, noise addition, or sampling, ignoring actual observations. 
Observational datasets in real-world are proxies without direct link to the target states. For instance, operational forecasting models may assimilate diverse data sources like weather station outputs, satellite imagery, and LiDAR-derived point clouds.
Moreover, these approaches tend to rely on computer vision architectures that favor uniform resolution with limited multi-scale processing capabilities, neglecting rich observations with heterogeneous spatiotemporal resolutions. To address these problems, we then enrich DAMBench with multi-modal real-world observations. We also conducted quantities of experiments to demonstrate the positive effectiveness of such multi-modal information.

\subsection{Additional Multi-Modal Real-World Observations}
 As shown in Figure~\ref{fig:observation_components}, DAMBench rasterizes the Earth, integrates various data sources, and includes over 6820 billion daily Earth meteorological observation data collected globally within raster grids from 1979 to 2024, as well as spatial multi-modal geographic information supplements for all regions within global raster grids, including text descriptions and geographic images, aiming to advance spatio-temporal data analysis and spatial intelligence research.

\subsubsection{Station-based Observations}
The precipitation observations are collected from a globally distributed network of rain gauges maintained by the NOAA Climate Prediction Center (CPC), comprising over 16,000 stations worldwide. These gauges provide direct, high-fidelity measurements of daily rainfall, making them indispensable for long-term climatological analyses and model validation. Compared to satellite or radar estimates, gauge data serve as ground-truth references and are critical for calibrating indirect precipitation products. 
All data are resampled to a $121  \times 240$ grids to be aligned with reanalysis data.
This gridded precipitation fields serve as a high-quality input for data assimilation and model validation in DAMBench.

\subsubsection{Satellite-based Observations}
Satellite-based outgoing longwave radiation (OLR) observations are derived from NOAA polar-orbiting satellites, including NOAA-14,
NOAA-16, and NOAA-18 satellites~\citep{liebmann1996description, qu2024slam} and provide critical insights into Earth’s radiation budget and deep tropical convection. 
OLR observations effectively retain large-scale variability while minimizing artificial noise, enabling their integration into DA pipelines as a dense, gridded source of satellite-based radiative information. All data are resampled to a $121  \times 240$ grids to be aligned with reanalysis data.

\subsection{Multi-Modal Representation Adapter}

\begin{figure}
    \centering
    \includegraphics[width=\linewidth]{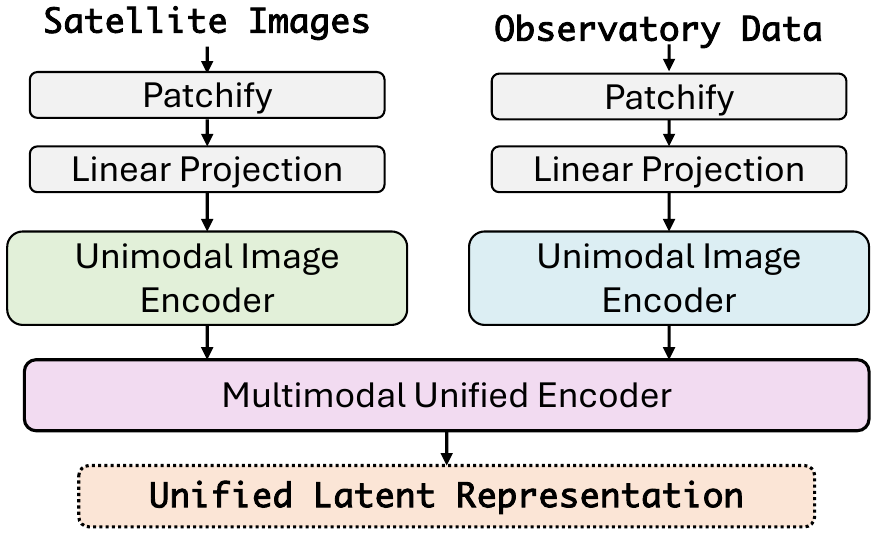}
    \caption{Overall framework of multi-modal representation adapter, a lightweight plugin for the integration of multi-modal observations.}
    \label{fig:plugin}
\end{figure}

To fully leverage multi-modal observational information, we introduce a lightweight yet effective plugin module that can be seamlessly integrated into existing data assimilation pipelines. As illustrated in Figure~\ref{fig:plugin}, satellite images and precipitation station data are first processed independently using identical encoders, before being fused via a shared multi-modal encoder into a unified latent representation.

For clarity, we detail the processing of satellite imagery $I$; the same architecture is applied to the precipitation observations $R$.

Each modality is divided into a sequence of non-overlapping patches. For satellite images, we denote the patches as $I_{p}$, which are then linearly embedded into a dense representation: 
$ \boldsymbol{e}_{p}^{I} = \textbf{W}_{p}^I {I}_{p}^{\top} + {b}_{p}^I$, 
where $\textbf{W}_{p}^I$ and and ${b}_{p}^I$ are learnable projection parameters. The learnable positional embeddings $\textbf{E}_I$  are further added to provide information about the relative position of each patch: $ \boldsymbol{e}_{E}^{I}=\boldsymbol{e}_{p}^{I}+\textbf{E}_I$.
Then, $\boldsymbol{e}_{PE}^{I}$ is sent to the layers of the self-attention module to integrate the sequence information: 
\begin{equation}\label{eq:MSA1}
    \left(\mathbf{Q}^{I}, \mathbf{K}^{I}, \mathbf{V}^{I} \right)^{\top} =\boldsymbol{e}_{PE}^{I} \left(\mathbf{W}_Q, \mathbf{W}_K, \mathbf{W}_{V}\right)^{\top},
\end{equation}
where $\textbf{W}_Q, \textbf{W}_K$, and $\textbf{W}_V$ are learnable matrices. The single-head and multi-head self attention ($\operatorname{Attn}(\cdot)$) are defined as: 
\begin{equation}\label{eq:MSA2}
\begin{split}
    \boldsymbol{e}^I_{(i)}=\operatorname{Softmax}\left(\textbf{Q}^{I} \textbf{K}^{{I}^{\top}} / \sqrt{d}\right) \textbf{V}^{I},\\
    \boldsymbol{e}^I_{Attn}=\operatorname{Concat}(\boldsymbol{e}^I_{(1)},\boldsymbol{e}^I_{(2)},...,\boldsymbol{e}^I_{(H)})\textbf{W}_{O},
\end{split}
\end{equation}
where $H$ is the number of attention heads, $\textbf{W}_{O}$ is a learnable weight matrix, and $\operatorname{Concat}(\cdot)$ denotes the concatenation function. After residual connection and layer normalization, the satellite image representation becomes: 
\begin{equation}    \boldsymbol{e}^{I}=\operatorname{LayerNorm}\left(\boldsymbol{e}^{I}_{E}+\operatorname{Attn}\left(\boldsymbol{e}^{I}_{E}\right)\right).
\end{equation}
The same procedure is applied to precipitation observations $R$ to obtain $\boldsymbol{e}^R$.
To align information from both modalities, we concatenate the modality-specific representation and feed them into a shared transformer encoder:
$\boldsymbol{e}^{\text{fused}} = \operatorname{Concat}(\boldsymbol{e}^I, \boldsymbol{e}^R),$
$ \boldsymbol{z}^{\text{fused}} = \operatorname{TransformerEncoder}(\boldsymbol{e}^{\text{fused}}).$

This multi-modal unified encoder enables the extraction of unified latent representations $\boldsymbol{z}^{\text{fused}}$ that integrate complementary spatial and semantic cues from both satellite imagery and station-based data.
The resulting representation $\boldsymbol{z}^{\text{fused}}$ can be directly consumed by downstream forecasting or data assimilation modules, enhancing performance under realistic multi-modal settings with minimal architectural overhead.



\subsection{Experiments}

To comprehensively assess the impact of incorporating real-world multi-modal observations, we augment each baseline method using our proposed AutoEncoder plugin. Specifically, the fused multi-modal representation $\boldsymbol{z}^{\text{fused}}$ is projected into the same latent space as the original model-specific embeddings, allowing seamless integration with existing architectures.

Table~\ref{tab:multimodal_experiments} summarizes the performance improvements across all baseline models. Remarkably, the inclusion of multi-modal observations consistently improves performance, with average MSE reductions of approximately 5\% across most methods. This clearly highlights the value of integrating heterogeneous observation sources—such as satellite-derived OLR and ground-based precipitation—in complex real-world DA tasks.

Among all models, VAE-VAR~\citep{xiao2025vae} benefits the most from multi-modal input, achieving a notable 7.79\% relative improvement in MSE. We attribute this to the model’s variational architecture, which allows it to better capture the uncertainty and complementary nature of different modalities. The ability to learn richer latent representations under complex observation configurations proves particularly beneficial in posterior estimation.
In the case of score-based models, while SDA~\citep{rozet2023score} is limited by its fixed observation format, SLAM~\citep{qu2024slam} shows superior performance when guided by the multi-modal signal. The use of multi-modal inputs as conditional prompts improves both the score estimation during training and the diffusion trajectory during inference, further demonstrating the effectiveness of modality-aware generative modeling.

Overall, these findings reinforce the necessity of considering multi-modal observational data in DL-based DA frameworks. They also validate the flexibility of our plugin-based integration mechanism, which can be easily adapted to enhance a wide range of existing models without retraining them from scratch.

\section{Conclusion and Future Work}
This work introduces the DAMBench, a multi-modal real-world benchmark for atmospheric data assimilation task. DAMBench is a comprehensive benchmark encompassing various deep learning based method for atmospheric data assimilation spanning the earth, covering worldwide regions over the past 25 years. It includes real-world reanalysis data along with multi-modal observations including satellite images and precipitation data from observatory.
Based on a thorough introduction of the data sources and analysis of experimental results, we highlight the significant impact of the DAMBench on advancing data assimilation research and its potential for progressing towards general climate intelligence.

\bibliographystyle{ACM-Reference-Format}
\bibliography{sample-base}

\newpage
\appendix

\section{Existing Methods for Data Assimilation}
\label{sec:baselines}

To comprehensively evaluate the effectiveness of our proposed method, we benchmark against six recent and representative data assimilation models. These baselines cover a broad range of methodologies—from probabilistic generative modeling and latent neural processes to variational learning and structured AI pipelines. All of them have demonstrated strong empirical performance in either synthetic settings or real-world Earth system tasks.

\paragraph{FNP: Fourier Neural Processes for Arbitrary-Resolution Data Assimilation}

FNP~\citep{chen2024fnp} is designed to overcome the resolution limitations in traditional AI-based data assimilation frameworks. It builds on neural processes by integrating Fourier-based embeddings to extract spatial patterns across varying grid scales. FNP can natively ingest and assimilate observations at arbitrary resolutions without interpolation or resampling. Once trained on a fixed resolution, it generalizes to out-of-domain resolutions and observation densities with no additional fine-tuning. FNP’s flexibility is particularly useful for operational settings where sensors report data on diverse spatial grids or in irregular formats.

\paragraph{VAE-Var: Variational Autoencoder-Enhanced Variational Assimilation in Meteorology}

VAE-Var~\citep{xiao2025vae} blends deep generative modeling with classical variational data assimilation. It introduces a variational autoencoder to learn a more expressive, non-Gaussian background distribution $p(x \mid x_b)$, alleviating the strong Gaussian assumption in conventional 3D-Var and 4D-Var methods. VAE-Var retains the standard variational objective, minimizing a composite cost comprising observation and background terms, but with a learned decoder for flexible optimization:
\[
\mathcal{L}(x) = -\log p(y \mid x) - \log p(x \mid x_b).
\]
Its structure-aware design enables compatibility with irregular observations and arbitrary observation operators. Experimental results on real-world weather forecasting data show that VAE-Var outperforms traditional methods and other neural baselines under sparse observation conditions.

\paragraph{SDA: Score-based Data Assimilation}

SDA~\citep{rozet2023score} proposes a principled, generative approach to trajectory-level data assimilation by leveraging continuous-time score-based diffusion. A score model is trained on short trajectory segments, enabling the non-autoregressive generation of full state trajectories via reverse SDE sampling. Notably, SDA decouples the observation model from training, using it only at inference time. This design enables zero-shot generalization to new observation patterns and flexible conditioning under different data availability scenarios. Its theoretical grounding and full-trajectory modeling set it apart from prior point-wise or filtering-based methods.

\paragraph{SLAM: Deep Generative Data Assimilation in Multimodal Setting}

SLAM~\citep{qu2024deep} is a diffusion-based data assimilation framework designed to operate in a unified latent space. It embeds multimodal observations (e.g., satellite images, in-situ sensor readings) and background states into a shared latent representation, enabling conditional generation via score-based diffusion. By modeling data assimilation as a posterior sampling problem, SLAMS effectively bypasses the need for explicitly defined observation operators. It supports zero-shot conditioning by leveraging a Bayesian decomposition of the score:
\[
\nabla_{x(t)} \log p(x(t) \mid y) = \nabla_{x(t)} \log p(x(t)) + \nabla_{x(t)} \log p(y \mid x(t)).
\]
Extensive ablations show that SLAMS is robust in noisy, sparse, and multi-resolution observation scenarios, and is especially effective for tasks like vertical temperature profile calibration in global climate models.

\paragraph{Adas: Towards an end-to-end artificial intelligence driven global weather forecasting system}

Adas~\citep{chen2023adas} represents a significant step toward end-to-end AI-based numerical weather prediction. It incorporates domain knowledge from traditional DA (e.g., confidence weighting) into a gated neural network architecture. Adas employs gated convolutions and gated cross-attention, modulated by a confidence matrix, to fuse sparse observations with background states. It is designed to operate in tandem with advanced AI forecasters like FengWu and has been shown to outperform the ECMWF Integrated Forecasting System (IFS) in long-range forecast skill for the first time. Adas is notable for its real-world deployment feasibility and strong performance under varying observation densities and qualities.

\paragraph{ConvCNP: Convolutional Conditional Neural Processes}

ConvCNP~\citep{gordon2019convcnp} belongs to the neural process family and introduces translation-equivariant functional embeddings to support structured prediction over spatial domains. It models the prediction function as a continuous convolutional map over a discretized input space, enabling flexible and uncertainty-aware interpolation across irregular or sparse observation contexts. Unlike traditional models tied to fixed grids, ConvCNP handles off-grid observations and allows zero-shot prediction across domains. Its architectural simplicity and principled inductive biases make it a competitive baseline for spatiotemporal field completion and assimilation tasks.

These existing approaches reflect the current frontier of data assimilation research across multiple paradigms—variational inference (VAE-Var), neural processes (FNP, ConvCNP), diffusion models (SLAMS, SDA), and hybrid attention-based architectures (Adas). Together, they form a rigorous benchmark for evaluating the flexibility, generalizability, and scientific robustness of new data assimilation models.

\begin{table*}[htbp]
\centering
\caption{Comparison between Traditional DA Methods and Deep Learning Approaches}
\label{tab:da_comparison}
\resizebox{0.98\linewidth}{!}
{\begin{tabular}{lcccc}
\toprule
\textbf{Method} & \textbf{Nonlinearity Handling} & \textbf{Uncertainty Estimation} & \textbf{Physical Consistency} & \textbf{Computational Cost} \\
\midrule
4D-Var & Limited (adjoint-based) & Point Estimate & Strong (via PDE constraints) & High \\
EnKF & Moderate (Gaussian, linear update) & Ensemble-based covariance & Moderate & Medium \\
Particle Filter & Strong (non-Gaussian, multimodal) & Full posterior (weighted particles) & Weak & Very High \\
Weak-Constraint 4D-Var & Moderate (linearized model error) & Point Estimate with model error & Improved & Very High \\
\midrule
DL Methods  & Strong & Generative or probabilistic modeling & Weak or Implicit & Low (after training) \\
\bottomrule
\end{tabular}}
\end{table*}

\section{Introduction to Traditional Data Assimilation Methods}

Traditional data assimilation (DA) aims to optimally combine a prior model predictions and noisy, sparse observations to estimate the evolving state of a dynamical system. Classical DA techniques are grounded in Bayesian estimation theory and the physics of the underlying model, with widespread use in meteorology, oceanography, and hydrology. In this section, we provide a mathematical overview of several foundational DA methods: Four-Dimensional Variational Assimilation (4D-Var), the Ensemble Kalman Filter (EnKF), the Particle Filter (PF), and the Weak-Constraint 4D-Var.

\subsection{Four-Dimensional Variational Method (4D-Var)}

4D-Var casts the assimilation problem as a trajectory optimization problem over a time window $[t_0, t_L]$. Given a nonlinear forecast model $\mathcal{M}_{0 \rightarrow l}$, a background state $\mathbf{x}_b$, and a sequence of observations $\{\mathbf{y}_1, \dots, \mathbf{y}_L\}$, the method seeks an initial state $\mathbf{x}_0$ that minimizes the cost function:

\begin{equation}
\begin{aligned}
&\min_{\mathbf{x}_0} J(\mathbf{x}_0) = \ \frac{1}{2} (\mathbf{x}_0 - \mathbf{x}_b)^\top \mathbf{B}^{-1} (\mathbf{x}_0 - \mathbf{x}_b) \\
& + \frac{1}{2} \sum_{l=1}^L \left( \mathbf{y}_l - \mathcal{H}_l(\mathcal{M}_{0 \rightarrow l}(\mathbf{x}_0)) \right)^\top \mathbf{R}_l^{-1} \left( \mathbf{y}_l - \mathcal{H}_l(\mathcal{M}_{0 \rightarrow l}(\mathbf{x}_0)) \right)
\end{aligned}
\end{equation}

Here, $\mathbf{B}$ and $\mathbf{R}_l$ denote the background and observation error covariance matrices, respectively, and $\mathcal{H}_l$ is the nonlinear observation operator at time $t_l$. The optimization is typically performed using adjoint-based gradient descent methods. 4D-Var ensures dynamical consistency through explicit model integration, but the computational burden is high due to the need for repeated forward and backward model runs. Additionally, it provides only a point estimate of the state rather than a full posterior distribution.

\subsection{Ensemble Kalman Filter (EnKF)}

The Ensemble Kalman Filter (EnKF) is a Monte Carlo approximation of the Kalman filter suitable for nonlinear, high-dimensional systems. It evolves an ensemble of $N_e$ model states $\{\mathbf{x}_k^{(n)}\}_{n=1}^{N_e}$ to estimate both the forecast and analysis error statistics. The forecast step propagates each ensemble member:

\begin{equation}
\mathbf{x}_k^{f(n)} = \mathcal{M}_{k-1 \rightarrow k}(\mathbf{x}_{k-1}^{a(n)})
\end{equation}

The analysis step updates the forecast with perturbed observations $\mathbf{y}_k^{(n)}$:

\begin{equation}
\mathbf{x}_k^{a(n)} = \mathbf{x}_k^{f(n)} + \mathbf{K}_k \left( \mathbf{y}_k^{(n)} - \mathcal{H}_k(\mathbf{x}_k^{f(n)}) \right)
\end{equation}

The Kalman gain $\mathbf{K}_k$ is computed using the ensemble sample covariance:

\begin{equation}
\mathbf{K}_k = \mathbf{P}_k^f \mathcal{H}_k^\top \left( \mathcal{H}_k \mathbf{P}_k^f \mathcal{H}_k^\top + \mathbf{R}_k \right)^{-1}
\end{equation}

\begin{equation}
\mathbf{P}_k^f = \frac{1}{N_e - 1} \sum_{n=1}^{N_e} \left( \mathbf{x}_k^{f(n)} - \bar{\mathbf{x}}_k^f \right) \left( \mathbf{x}_k^{f(n)} - \bar{\mathbf{x}}_k^f \right)^\top
\end{equation}

EnKF provides a flow-dependent estimate of the error covariance, allowing adaptation to the evolving dynamics. However, the method assumes approximate Gaussianity and linearity in the update, which limits performance in strongly nonlinear or multimodal settings. Techniques such as covariance localization and inflation are often employed to mitigate sampling errors due to limited ensemble size.

\subsection{Particle Filter (PF)}

The Particle Filter (PF), or Sequential Monte Carlo method, seeks to approximate the full posterior distribution using a set of weighted particles $\{ \mathbf{x}_k^{(n)}, w_k^{(n)} \}_{n=1}^{N_p}$. The forecast step samples from the dynamical model:

\begin{equation}
\mathbf{x}_k^{(n)} \sim p(\mathbf{x}_k | \mathbf{x}_{k-1}^{(n)})
\end{equation}

The weights are updated using the likelihood of the observation:

\begin{equation}
w_k^{(n)} \propto w_{k-1}^{(n)} \cdot p(\mathbf{y}_k | \mathbf{x}_k^{(n)})
\end{equation}

Particles are resampled when the effective sample size falls below a threshold, to avoid degeneracy. Unlike EnKF, the particle filter can represent arbitrary posterior shapes and is exact in the limit of infinite particles. However, in high-dimensional systems, it suffers from the curse of dimensionality—most particles obtain negligible weight, leading to severe sample impoverishment.

\subsection{Weak-Constraint 4D-Var}

The standard 4D-Var assumes a perfect model (strong constraint). The Weak-Constraint 4D-Var relaxes this assumption by explicitly incorporating model error $\mathbf{q}_l$:

\begin{equation}
\mathbf{x}_{l+1} = \mathcal{M}_l(\mathbf{x}_l) + \mathbf{q}_l
\end{equation}

The cost function is modified to penalize both background and model errors:

\begin{equation}
\begin{aligned}
J(\mathbf{x}_0, \{\mathbf{q}_l\}) =\ & \frac{1}{2} (\mathbf{x}_0 - \mathbf{x}_b)^\top \mathbf{B}^{-1} (\mathbf{x}_0 - \mathbf{x}_b) \\
& + \frac{1}{2} \sum_{l=1}^L \left( \mathbf{y}_l - \mathcal{H}_l(\mathbf{x}_l) \right)^\top \mathbf{R}_l^{-1} \left( \mathbf{y}_l - \mathcal{H}_l(\mathbf{x}_l) \right) \\
& + \frac{1}{2} \sum_{l=0}^{L-1} \mathbf{q}_l^\top \mathbf{Q}_l^{-1} \mathbf{q}_l
\end{aligned}
\end{equation}

Here, $\mathbf{Q}_l$ is the model error covariance. This formulation allows better handling of structural model deficiencies and long assimilation windows. However, it introduces significantly more degrees of freedom, making the optimization more computationally intensive and sensitive to assumptions about model error statistics.

\subsection{Methods Comparison}
Table~\ref{tab:da_comparison} provides a comparative summary of traditional data assimilation methods and modern deep learning-based approaches. While traditional methods maintain strong physical consistency and interpretability, they are often computationally intensive and limited in handling nonlinear or high-dimensional uncertainty. In contrast, deep learning methods provide flexible nonlinear modeling and scalable inference, but often lack explicit enforcement of physical laws unless guided by hybrid or physics-informed designs.

In recent years, deep learning (DL) approaches have emerged as a promising alternative or complement to traditional data assimilation (DA) techniques. These models, such as neural operators, diffusion models, and transformer-based architectures, are capable of capturing complex nonlinear relationships in high-dimensional geophysical systems, often with significantly reduced inference cost after training. Unlike traditional methods that rely on explicit physical models and linearized operators, DL-based methods implicitly learn mappings from background states and observations to analyses, enabling data-driven generalization across variables, resolutions, and spatiotemporal scales.

Despite these advantages, DL approaches face several challenges. First, their ability to estimate uncertainty varies across architectures; while probabilistic generative models (e.g., diffusion or flow-based models) provide uncertainty quantification, deterministic models typically lack this feature. Second, pure DL methods often struggle to enforce physical constraints such as conservation laws or dynamic consistency unless augmented by physics-informed loss terms or hybrid training. Third, while traditional methods offer theoretical guarantees under Gaussian assumptions, DL models are more opaque and require extensive validation, especially in safety-critical applications like numerical weather prediction.

To address these limitations, recent efforts have explored physics-constrained deep learning, variational surrogate models, and neural process families that integrate structural priors or PDE operators into the learning pipeline. These hybrid designs aim to combine the scalability and expressiveness of DL with the robustness and interpretability of physics-based assimilation frameworks.

\section{Implementation Details}

\subsection{Datasets}

ERA5 is the fifth generation of atmospheric reanalysis produced by the European Centre for Medium-Range Weather Forecasts (ECMWF) as part of the Copernicus Climate Change Service (C3S). It provides a comprehensive global record of atmospheric, land-surface, and sea-state variables dating from 1940 to the present, with hourly temporal resolution and a spatial resolution of approximately 31~km (0.25°). ERA5 replaces the earlier ERA-Interim dataset and incorporates a significantly upgraded data assimilation system (IFS Cycle 41r2) with a higher-resolution model and enhanced observation ingestion.

The dataset is generated using a 4D-Var data assimilation system that combines model forecasts with a wide array of historical observations, including satellite radiances, radiosondes, surface stations, aircraft reports, and buoy data. This integration results in physically consistent estimates of various climate variables such as temperature, wind, humidity, precipitation, and surface pressure across multiple vertical levels.

ERA5 is widely used in climate research, weather forecasting, hydrology, renewable energy assessments, and machine learning for Earth system modeling due to its high spatiotemporal resolution, long historical coverage, and consistent quality. In particular, its hourly data granularity makes it especially suitable for training data-driven models that require fine-scale temporal dynamics and accurate ground-truth representations.
\begin{table}[ht]
    \centering
    \begin{tabular}{ccc}
    \toprule
         Name  &Description &Level\\
         \midrule
          u10 &x-direction wind at 10m height &Single\\
          v10 &y-direction wind at 10m height &Single\\
          t2m &Temperature at 2m height &Single\\
          mslp  & mean sea level pressure &Single\\
          z   &Geopotential & 13\\
          q   &Specific humidity &13\\
          u   &x-direction wind &13\\
          v   &y-direction wind &13\\
          T   &Temperature &13\\

    \bottomrule
    \end{tabular}
   \caption{Atmospheric Variables Considered. The 13 levels are 50, 100, 150, 200, 250, 300, 400, 500, 600, 700, 850, 925, 1000 hPa.}
    \label{tab:dataset}
\end{table}

\subsection{Experimental Setup}
We train the models using the AdamW optimizer, which decouples weight decay from the gradient update, providing improved generalization performance in modern deep learning settings. The learning rate is scheduled using a warm-up cosine annealing strategy: the training begins with a learning rate of $10^{-5}$, which linearly increases and reaches a peak of $10^{-4}$ after the first $\frac{1}{6}$ of the total training steps. It then gradually decays following a cosine curve down to a final value of $3 \times 10^{-6}$, encouraging smoother convergence in the later training phases.

Training is performed using data-parallelism across 8 NVIDIA A800 GPUs. We adopt a global batch size of 8, with each GPU processing one sample per iteration. The model is trained for 40 full epochs on the training dataset, which corresponds to approximately two days of wall-clock time. Mixed precision training is enabled using NVIDIA's Apex or PyTorch AMP to accelerate computation and reduce memory usage.

During inference, we run the model on a single A800 GPU node. For each forecast window, the model generates a fully assimilated state, including all predicted variables, in approximately 15 minutes. This includes both the encoding of observations and the generation of posterior predictions. The inference process is deterministic, with dropout and stochastic augmentations disabled to ensure stable and reproducible outputs.

All experiments are conducted under a fixed random seed for model initialization and data shuffling to ensure reproducibility. Unless otherwise stated, all results reported in the main text are averaged over three independent runs to account for variability due to training stochasticity.

\subsection{Evaluation Metrics}

We evaluate the performance of our data assimilation models using a comprehensive suite of metrics that measure both numerical accuracy and physical fidelity. Specifically, we adopt:

\begin{itemize}
    \item \textbf{Pointwise Accuracy Metrics:} Mean Absolute Error (MAE), Mean Squared Error (MSE), and Latitude-weighted Root Mean Squared Error (RMSE);
    \item \textbf{Spectral Fidelity Metrics:} Spectral Divergence (SpecDiv) and Spectral Residual (SpecRes);
\end{itemize}

Together, these metrics quantify discrepancies between the predicted state $\hat{x}_{h,w,c}$ and the reference truth $x_{h,w,c}$ across both spatial and spectral domains.

\paragraph{Mean Absolute Error (MAE) and Mean Squared Error (MSE).}
MAE and MSE capture average pointwise differences, and are defined as:

\begin{equation}
\operatorname{MAE} = \frac{1}{HWC} \sum_{c=1}^{C} \sum_{h=1}^{H} \sum_{w=1}^{W} \left| x_{h,w,c} - \hat{x}_{h,w,c} \right|,
\end{equation}

\begin{equation}
\operatorname{MSE} = \frac{1}{HWC} \sum_{c=1}^{C} \sum_{h=1}^{H} \sum_{w=1}^{W} \left( x_{h,w,c} - \hat{x}_{h,w,c} \right)^2.
\end{equation}

While MAE is more robust to outliers and provides a direct measure of absolute error, MSE penalizes large deviations more heavily due to the squaring, and is commonly used in optimization objectives.

\paragraph{Latitude-weighted Root Mean Squared Error (RMSE).}
Since Earth is a sphere, grid points near the equator represent larger physical areas than those near the poles. To correct for this distortion, we apply a latitude-based weighting using $\cos(\alpha_{h})$, where $\alpha_{h}$ is the latitude (in radians) at row $h$. The per-channel weighted RMSE is defined as:

\begin{equation}
\operatorname{RMSE}(c) = \sqrt{ \sum_{h=1}^{H} \sum_{w=1}^{W} \omega_{h} \left( x_{h,w,c} - \hat{x}_{h,w,c} \right)^2 },
\end{equation}
where
\[
\omega_{h} = \frac{\cos(\alpha_{h})}{\sum_{h'=1}^{H} \cos(\alpha_{h'})} \cdot \frac{1}{W}.
\]

This weighting ensures spatial errors are representative of actual Earth surface area, as commonly adopted in geophysical forecast evaluation.

\paragraph{Physical Evaluation via Spectral Divergence (SpecDiv).}

The Spectral Divergence metric, denoted as $\mathcal{M}_{\mathrm{SpecDiv}}$, quantifies the difference between the predicted and ground-truth power spectral distributions in the Fourier domain. Inspired by the Kullback–Leibler (KL) divergence, this metric captures the relative entropy between two normalized spectral densities, emphasizing mismatches in energy allocation across spatial frequencies.

Given a predicted field $\hat{x}$ and a ground-truth field $x$, we compute their 2D Fourier transforms to obtain spectral energy densities $\hat{S}(k)$ and $S(k)$, where $k \in \mathbf{K}$ is the scalar wavenumber. We then restrict the comparison to high-frequency components using a quantile-based filter:
\[
\mathbf{K}_q = \{k \in \mathbf{K} \mid k \geq Q(q)\}, \quad q \in [0, 1]
\]
where $Q(q)$ denotes the $q$-th quantile of all scalar wavenumbers.

For each wavenumber $k \in \mathbf{K}_q$, we normalize the power spectra:
\[
S'(k) = \frac{S(k)}{\sum_{k' \in \mathbf{K}_q} S(k')}, \quad \hat{S}'(k) = \frac{\hat{S}(k)}{\sum_{k' \in \mathbf{K}_q} \hat{S}(k')}
\]

The spectral divergence is then defined as:
\begin{equation}
\mathcal{M}_{\mathrm{SpecDiv}} = \sum_{k \in \mathbf{K}_q} S'(k) \cdot \log \left( \frac{S'(k)}{\hat{S}'(k)} \right)
\label{eq:specdiv}
\end{equation}

This metric evaluates how closely the predicted spectral energy distribution aligns with that of the reference. A lower $\mathcal{M}_{\mathrm{SpecDiv}}$ indicates better spectral fidelity, with $\mathcal{M}_{\mathrm{SpecDiv}} = 0$ when the two spectra are identical. Since high-frequency components are critical for preserving spatial details and physical structures, this metric provides a physics-aware signal fidelity measure that complements traditional pixel-wise metrics such as RMSE or SSIM.

Together, these metrics offer a comprehensive view of model performance, covering pointwise accuracy (MAE, MSE), geospatial realism (latitude-weighted RMSE), and physical consistency (SpecDiv) in high-dimensional data assimilation tasks.

\section{Recent Explorations for General Spatio-Temporal Intelligence}

As the scope of environmental data assimilation expands beyond single-variable or single-domain tasks, the research community has increasingly turned its attention to General Spatio-Temporal Intelligence (GSTI), a paradigm aimed at learning universal representations that can generalize across multiple regions, scales, modalities, and tasks~\citep{jindong2021air, wang2025language}.

Recent advancements have leveraged large-scale deep models, including graph neural networks, neural operators, and vision transformers, to encode multivariate physical fields such as temperature, pressure, wind, and precipitation. These efforts often build on benchmarks like WeatherBench2~\citep{rasp2024weatherbench}, which enable standardized evaluation across diverse spatiotemporal prediction settings.

A key direction in GSTI is the development of foundation models for physical systems, exemplified by models like GraphCast, Pangu-Weather, and FourCastNet~\citep{lam2023graphcast, bi2023pangu, fourcastnet}. These models are trained on decades of reanalysis data and are capable of making medium-range forecasts at global scale with remarkable accuracy and efficiency. Their success stems from architectural innovations (e.g., Fourier neural operators, long-range self-attention) and scale—often involving billions of parameters and petascale data.

In parallel, hybrid physics-AI approaches have also gained traction, aiming to embed physical constraints, conservation laws, or explicit partial differential equations (PDEs) into neural architectures~\citep{raissi2019physics, wang2025phyda}. For instance, some models use physics-informed loss functions, while others directly learn residuals or corrections to numerical solvers, achieving better interpretability and robustness.

Moreover, the emergence of multi-modal data sources—including satellite imagery, ground station readings, radar maps, and ocean buoys—has motivated the design of multi-source fusion architectures. These architectures encode each modality independently and then learn cross-modal relationships, often using attention or shared latent spaces, to improve assimilation fidelity.

Finally, the integration of generative modeling, such as diffusion-based score models and flow-matching networks, has shown promise in capturing uncertainty and producing physically plausible state reconstructions, particularly under sparse or noisy observational regimes.

Overall, these advances signal a shift from narrow-task optimization to more general-purpose, transferable intelligence over complex geophysical systems.

\section{Broader Impact}

This work introduces DAMBench, a unified and realistic benchmark for deep learning-based atmospheric data assimilation (DA). By integrating large-scale, multi-modal, and real-world observations, our benchmark aims to accelerate the development of scalable and generalizable DA algorithms that are vital for modern Earth system modeling. The broader impact of this work spans several dimensions:

\textbf{Scientific Advancement.} DAMBench facilitates reproducible and rigorous evaluation of learning-based DA methods, addressing a longstanding gap in the field. This can foster collaboration between the climate science and machine learning communities, catalyzing innovations in hybrid physical–AI modeling, uncertainty-aware inference, and generative forecasting.

\textbf{Environmental and Societal Applications.} Improved data assimilation models can directly benefit high-stakes applications such as weather forecasting, climate change mitigation, and disaster response. Enhanced accuracy and timeliness in assimilating diverse observations (e.g., from satellite and ground sensors) can support better early warning systems for extreme weather events and long-term climate resilience planning.

\textbf{Fairness and Accessibility.} DAMBench is open-source and constructed from publicly available data, ensuring accessibility for researchers worldwide, including those with limited computational resources. By supporting multi-resolution and modular evaluation, it enables inclusive participation from both domain experts and ML practitioners at various levels.

We hope this benchmark will serve not only as a foundation for robust scientific inquiry but also as a catalyst for impactful real-world applications aligned with the goals of environmental sustainability and technological equity.

\section{Discussion}

Our study introduces DAMBench, a unified and realistic benchmark tailored for evaluating deep learning-based data assimilation (DA) systems under operational-like conditions. Through rigorous experimental design and comprehensive comparisons across representative methods—including latent generative models, neural process frameworks, and diffusion-based approaches—we provide the first large-scale, multi-modal assessment of DA performance grounded in real-world atmospheric data.

The results highlight several key insights. First, deep learning models significantly improve over background fields, affirming their capacity to reconstruct high-dimensional system states from sparse observations. Notably, methods like FNP and VAE-VAR demonstrate superior accuracy and physical consistency, showcasing the effectiveness of structured uncertainty modeling and variational inference. Second, our proposed multi-modal plugin further amplifies performance, underscoring the critical role of heterogeneous observations in real-world DA. These findings suggest that integrating diverse sensing modalities—not commonly addressed in prior work—can yield substantial benefits in assimilation accuracy and robustness.

Importantly, DAMBench bridges a long-standing gap between algorithmic innovation and realistic DA settings. Prior studies often relied on synthetic or overly simplified setups, making it difficult to compare methods fairly or assess their generalizability. By contrast, DAMBench provides structured observation-background-target triplets, a unified evaluation protocol, and a scalable data pipeline compatible with large-scale DL training.

Looking ahead, DAMBench opens several avenues for future research. It provides a foundation for testing emerging paradigms such as foundation models for Earth science, probabilistic assimilation via diffusion or score-based methods, and adaptive multi-modal fusion. It also invites the community to explore efficiency-accuracy trade-offs, long-term forecast stability post-assimilation, and uncertainty quantification under different observation regimes.

We hope that DAMBench will serve as a catalyst for systematic, reproducible, and forward-looking research in learning-based data assimilation, accelerating progress toward operational integration of AI into next-generation Earth system modeling.




\end{document}